\pdfoutput=1

\documentclass[11pt]{article}

\usepackage{acl}

\usepackage{times}
\usepackage{latexsym}

\usepackage{microtype}
\usepackage{amsmath}
\usepackage{stfloats}
\usepackage{microtype}
\usepackage{balance}
\usepackage{paralist}
\usepackage{verbatim}
\usepackage[normalem]{ulem}
\usepackage[vlined,ruled,linesnumbered]{algorithm2e}
\usepackage{makecell}
\usepackage{txfonts}
\usepackage{xspace}
\usepackage{booktabs}
\usepackage{graphicx}
\usepackage{color,soul}
\usepackage{xcolor, colortbl}
\usepackage{multirow}
\usepackage{booktabs}
\usepackage{subcaption}
\usepackage{enumitem}
\usepackage[toc,page]{appendix}

\usepackage[T1]{fontenc}

\newcommand\sect[1]{\S\ref{#1}}

\newcommand{\revised}[1]{{\color{blue}#1}\xspace}
\renewcommand{\revised}[1]{#1\xspace}

\newcommand{\note}[2]{\xspace{\color{#1}{[#2]}}\xspace}

\newcommand{\hao}[1]{\note{orange}{Hao: #1}}

\newcommand{\sysname}{\textsc{Tailor}\xspace}
\newcommand{\repourl}{\url{https://github.com/allenai/tailor}}

\newcommand{\polyjuice}{{\textsc{Polyjuice}}\xspace}

\newcommand{\gptbaseline}{\textsc{GPT-2}\xspace}
\newcommand{\tfive}{\textsc{T5}\xspace}

\newcommand{\retrieveedit}{\textsc{RetrieveEdit}\xspace}
\newcommand{\csgpttv}{\textsc{CS-GPT-TV}\xspace}
\newcommand{\csgpttp}{\textsc{CS-GPT-TP}\xspace}
\newcommand{\csgpt}{\textsc{CS-GPT}\xspace}
\newcommand{\csgptzero}{\textsc{CS-Sys-Gen}\xspace}
\newcommand{\styleptb}{\textsc{StylePTB}\xspace}

\newcommand{\eg}{\emph{e.g.,}\xspace}%
\newcommand{\ie}{\emph{i.e.,}\xspace}

\newcommand{\colbox}[2]{\colorbox{#1}{#2}}
\newcommand{\veryshortarrow}[1][3pt]{\mathrel{%
   \hbox{\rule[\dimexpr\fontdimen22\textfont2-.2pt\relax]{#1}{.4pt}}%
   \mkern-4mu\hbox{\usefont{U}{lasy}{m}{n}\symbol{41}}}}
\definecolor{caddback}{HTML}{e5faf5}
\definecolor{cadd}{HTML}{007857}
\definecolor{cdelback}{HTML}{ffefea}
\definecolor{cdel}{HTML}{d35229}
\def \arrow{$\veryshortarrow$}
\newcommand{\add}[1]{\colbox{caddback}{\color{cadd}#1\xspace}} %

\newcommand{\remove}[1]{\colbox{cdelback}{{\color{cdel}#1\xspace}}}%
\newcommand{\removesout}[1]{\colbox{cdelback}{{\color{cdel}\sout{#1}\xspace}}}%

\newcommand{\swap}[2]{\remove{#1}\arrow\add{#2}}

\definecolor{clightgrey}{HTML}{80868b}
\definecolor{clabel}{HTML}{3B4D73}
\definecolor{csparsity}{HTML}{174ea6}

\newcommand{\tblank}[1]{\texttt{<id\_#1>}\xspace}

\newcommand{\tcolored}[2]{#2}
\newcommand{\tctrl}[2]{\texttt{\tcolored{#1}{#2}}}

\newcommand{\tlabel}[1]{\tctrl{clabel}{#1}}
\newcommand{\tsparsity}[1]{\tctrl{csparsity}{#1}}
\newcommand{\ttense}[1]{\tctrl{csparsity}{#1}}
\newcommand{\tvoice}[1]{\tctrl{csparsity}{#1}}
\newcommand{\tlemma}[1]{\tcolored{black}{#1}}

\newcommand{\theader}[3]{\tlabel{#1}+\tsparsity{#2}: \tlemma{#3}}
\newcommand{\theaderrandom}[2]{\tlabel{#1}: \tlemma{#2}}
\newcommand{\theaderVerb}[3]{\tlabel{VERB}+\tvoice{#1}+\ttense{#2}: \tlemma{#3}}

\newcommand{\tgenerateLabel}[1]{\texttt{\color{clightgrey}#1}: }
\newcommand{\tgenerate}[2]{{\color{clightgrey}{[\tgenerateLabel{#1}{\color{black}\tlemma{#2}}}]}\xspace}

\definecolor{cop}{HTML}{3B4D73}
\newcommand{\optag}[1]{\textcolor{cop}{\texttt{#1}}\xspace}

\title{\sysname: Generating and Perturbing Text with Semantic Controls}

\renewcommand{\textsuperscript}[1]{\raisebox{0.5ex}{#1}}
\renewcommand{\quad}{\hspace{0.5em}}

\author{Alexis Ross\thanks{\ \ denotes equal contribution.}\textsuperscript{$*\dagger$} \quad Tongshuang Wu\textsuperscript{$*\diamondsuit$} \quad Hao Peng\textsuperscript{$\diamondsuit$} \quad Matthew E.\ Peters\textsuperscript{$\dagger$} \quad Matt Gardner\textsuperscript{$\spadesuit$}\thanks{\ \ Work done while at Allen Institute for AI} \\ 
\textsuperscript{$\dagger$}Allen Institute for Artificial Intelligence, Seattle, WA, USA\\
\textsuperscript{$\diamondsuit$}Paul G.\ Allen School of Computer Science and Engineering, University of Washington \\
\textsuperscript{$\spadesuit$}Microsoft Semantic Machines, USA \\
\texttt{\{alexisr,matthewp\}@allenai.org} \\
\texttt{\{wtshuang,hapeng\}@cs.washington.edu} \\
{\texttt{mattgardner@microsoft.com}}
}

\begin{document}
\maketitle
\begin{abstract}

Controlled text perturbation is useful for evaluating \revised{and improving} model generalizability.
However, current techniques rely on training a model for every \revised{target perturbation},
which is expensive and hard to generalize.
We present \sysname, a semantically-controlled text generation system.
\sysname builds on a pretrained seq2seq model and produces textual outputs 
\revised{conditioned} on {control codes} derived from semantic representations.
We craft a set of operations to modify the control codes,
which in turn steer generation towards targeted attributes.
These operations can be further composed into higher-level ones, 
allowing for flexible perturbation strategies. \revised{We demonstrate the effectiveness of these perturbations in multiple applications.}
\revised{First,} we use \sysname to automatically create high-quality contrast sets for four distinct natural language processing (NLP) tasks.
These contrast sets contain fewer spurious 
\revised{artifacts}
and are complementary to manually annotated ones in \revised{their}
lexical diversity. 
\revised{Second}, we show that \sysname~\revised{perturbations can} improve model generalization through data augmentation. Perturbing just $\sim$2\% of training data \revised{leads to a} 5.8-point
gain on an NLI challenge set \revised{measuring reliance on syntactic heuristics}.
\end{abstract}

\section{Introduction}
\label{sec:introduction}

\begin{figure}[t]
\centering
\includegraphics[trim={0 9cm 30cm 0cm}, clip, width=1\columnwidth]{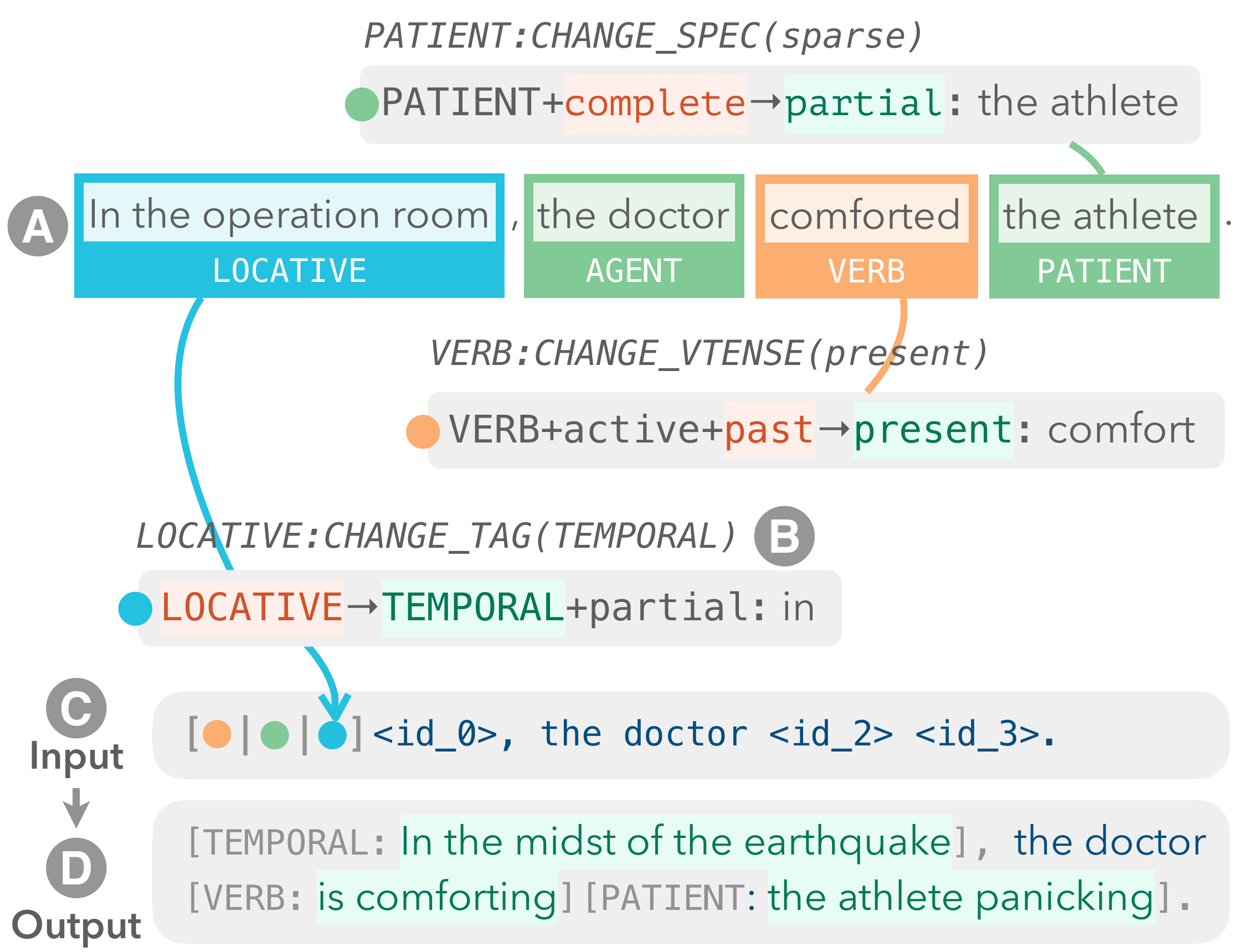}
\vspace{-20pt}
\caption{
A compositional perturbation using \sysname.\footnotemark ~Given (A) an original sentence, we abstract each span into a structured \emph{header} that contains its semantic roles and keywords. \revised{Arguments to preserve are included in the \emph{context}, along with \emph{blanks} (\texttt{<id\_*>}) denoting where new generated text may be inserted.} We specify desired perturbations by modifying each control code (\eg changing role \swap{\texttt{LOCATIVE}}{\texttt{TEMPORAL}} in (B), verb tense \swap{past}{present}, and patient keyword specificity \swap{complete}{partial}).
Given these \emph{perturbed control codes} in the input (C), \sysname generates a new sentence (D) that reflects the desired perturbations.
}
\vspace{-10pt}
\label{fig:teaser}
\end{figure}
\revised{Semantic perturbation through controlled text generation}
modifies sentences to match certain target attributes, such as verb tense or sentiment (\eg \emph{positive}$\rightarrow$\emph{negative}). 
It has been widely applied to a variety of tasks, \eg 
\revised{changing text style}~\cite{reid2021lewis}, 
mitigating dataset biases~\cite{gardner2021competency}, explaining model behaviors~\cite{ross2020explaining}, and improving model generalization~\cite{teney2020learning,polyjuice:acl21}.
Existing efforts train task-specific generators, 
\eg training a sentiment style transferer 
requires instances annotated with \emph{positive} and \emph{negative} labels~\cite{madaan2020generate}.
As a result, they require costly annotated data and re-training for every task of interest.

\footnotetext{We opensource \sysname and release \sysname-generated contrast sets at \repourl.}

This work introduces \sysname, a system that supports application-agnostic perturbations. 
At its core is a \emph{controlled generator} (\S\ref{sec:generator-design}) that flexibly generates outputs from target semantic attributes, \revised{which we represent through structured \textbf{control codes} in the inputs.}
As shown in Figure~\ref{fig:teaser}, \revised{these control codes} build on the PropBank semantic analysis \citep{palmer-etal-2005-proposition} 
of the original sentence: \revised{For each argument span, the \textit{semantic role} and 
\textit{keyword} control codes specify the desired semantic content for the span at varying levels of granularity.}
To encourage control code following,
we train \revised{the \sysname generator} with \textbf{unlikelihood training}~\cite{Welleck2020Neural} to penalize generations that are not aligned with designated \revised{control codes}.

The use of semantic role \revised{control codes} allows \sysname to perform fine-grained changes to individual arguments in a sentence (\eg one can change only the \optag{PATIENT} in Figure~\ref{fig:teaser}).
Instead of 
\revised{specifying a perturbation with a generic target property
(\eg \emph{positive}\arrow\emph{negative}), we can 
specify the linguistic transformation
used to achieve the property}
(\eg changing sentiment through negation or antonym replacement).
Making such fine-grained perturbations 
\revised{allows for more careful evaluation and improvement of}
models' language understanding~\cite{kaushik2019learning, polyjuice:acl21}. 

To highlight the perturbations facilitated by \sysname,
we craft a list of primitive \emph{perturbation operations} (\sect{sec:perturbations}) on inputs to the generator; these can be easily composed to achieve more complex perturbations.
In Figure~\ref{fig:teaser}, \sysname transforms sentence A to D through a series of perturbations:
syntactic rewriting (changing verb tense), then sentence expansion (extending ``the athlete''), and finally data recombination (\ie generating new text that contains ``in'' but follows the  \optag{TEMPORAL} control).
Compared to existing approaches that require training a separate model for every step or annotating a dataset that represents this transformation end-to-end, such compositions make \sysname more cost-effective and generalizable.
In fact, on nine fine-grained and compositional \textsc{StylePTB} perturbations~\citep{Lyu2021StylePTBAC}, \sysname achieves performance compatible with task-specific baselines, and even outperforms them on five transfers (\sect{sec:appendix-style-transfer}).

\sysname's flexible \revised{and human-readable} control codes allow for broad, easily extendable applicability.
We demonstrate its utility in evaluating and improving NLP model robustness, showing that \sysname can help replicate existing \textbf{contrast sets} on four diverse tasks.
By abstracting manual perturbation types in prior work into perturbation strategies with \sysname, we 
\revised{can apply}
the changes to larger datasets while saving manual annotation efforts.
Our analysis suggests that these contrast sets not only have high rates of validity, but also reduce spurious 
\revised{artifacts compared to the original evaluation datasets}.
In addition, \sysname-produced contrast sets complement human annotated ones in terms of lexical diversity: only $\sim$10\% of their unique tokens overlap with manually created contrast sets.
We also explore \sysname's utility in data augmentation. We find that augmenting training data with just
\revised{$\sim$2\% of \sysname perturbations} improves the \textbf{robustness} of natural language inference (NLI) models to inference heuristics, increasing performance on the HANS evaluation set~\cite{mccoy2019right} by an average of {5.81} points and outperforming a previous syntactic augmentation method for NLI.

\begin{table*}
\centering
\fontsize{8.5}{9.5}\selectfont
\renewcommand{\arraystretch}{0.9}
\setlength{\tabcolsep}{5pt}

\begin{tabular}{@{} c|p{0.48\linewidth}|p{0.31\linewidth}|p{0.13\linewidth} @{}}
\toprule
& \textbf{Input} & \textbf{Target Output} & \textbf{Description}\\
\midrule\midrule

\multirow{3}{*}{A} 
    & \multirow{3}{\linewidth}{\textcolor{cop}{[\theaderVerb{active}{past}{comfort} | \theader{AGENT}{complete}{the doctor} | \theader{PATIENT}{partial}{athlete} | \theader{LOCATIVE}{sparse}{*}]} \tblank{0}, \tblank{1} \tblank{2} \tblank{3}.}
    & [\tlabel{LOCATIVE}: In the operating room], [\tlabel{AGENT}: the doctor] [\tlabel{VERB}: comforted] [\tlabel{PATIENT}: the athlete]. 
    & \multirow{3}{\linewidth}{Mask all roles} \\\midrule

\multirow{2}{*}{B} 
    & \multirow{2}{\linewidth}{\textcolor{cop}{[\theaderVerb{active}{past}{comfort} | \theader{LOCATIVE}{sparse}{*}]} \tblank{0}, the doctor \tblank{1} \tblank{2} the athlete  \tblank{3}.} 
    & [\tlabel{LOCATIVE}: In the operating room], the doctor [\tlabel{VERB}: comforted] the athlete. 
    & \multirow{2}{\linewidth}{\emph{Empty} blanks}\\\midrule

\multirow{2}{*}{C} 
    & \multirow{2}{\linewidth}{\textcolor{cop}{[\theaderVerb{active}{past}{comfort} | \theader{LOCATIVE}{sparse}{*}]} \tblank{0}, the doctor \tblank{1} the athlete.} 
    & [\tlabel{LOCATIVE}: In the operating room], the doctor [\tlabel{VERB}: comforted] the athlete. 
    & \multirow{2}{\linewidth}{Mask \emph{subset} of arguments}\\\midrule\midrule
    
\multirow{3}{*}{\emph{N}} & \multirow{3}{\linewidth}{\textcolor{cop}{[\theaderVerb{\remove{passive}}{\remove{present}}{comfort} | \theader{\remove{PATIENT}}{complete}{the doctor} | \theader{\remove{AGENT}}{partial}{athlete} | \theader{\remove{TEMPORAL}}{sparse}{*}]} \tblank{0}, \tblank{1} \tblank{2} \tblank{3}.} 
& [\tlabel{TEMPORAL}: In the operating room], [\tlabel{PATIENT}: the doctor] [\tlabel{VERB}: comforted] [\tlabel{AGENT}: the athlete]. & \multirow{3}{\linewidth}{\emph{Negative} sample}\\

\bottomrule

\end{tabular}
\vspace{-5pt}
\caption{Example input/output formats for sentence ``In the operating room, the doctor comforted the athlete.'' 
A--C show different input formats the generator accepts. Each input (\sect{ssec:prompts}) contains a {\textbf{header} \revised{(in brackets)}, which contains \emph{control codes} (semantic role/keyword) for each span}, as well as a \textbf{context}, which includes \revised{both original text to preserve and} \emph{blanks} \revised{(\texttt{<id\_*>})} denoting where new text may be generated. 
\revised{The \sysname generator outputs text that infills the context's blanks with text following the header's control codes.} 
The last input (N) is a \emph{negative} sample used for unlikelihood training, as described in \sect{ssec:training}.}
\vspace{-10pt}
\label{tab:example-prompts}
\end{table*}

\section{\sysname's Controllable Generator}
\label{sec:generator-design}
Here, we provide an overview of the \sysname generator.
We first outline three types of \textbf{controls} (\sect{ssec:controls}) that allow for specifying sentence meanings at varying granularity. Next, we explain how to embed them within \textbf{inputs} (\sect{ssec:prompts}) to the generator. 
We train the generator to follow control codes with \textbf{unlikelihood training} (\sect{ssec:training}).

\subsection{Three Types of Controls}
\label{ssec:controls}
We use the following three types of controls to specify the shallow semantics, actual content, and ordering of various phrases in a sentence.

\textbf{Semantic roles} to denote shallow semantics.
We rely on the PropBank semantic formalism~\citep{ palmer-etal-2005-proposition}, as it provides well-established representations of meanings that are generalizable across different predicates and languages~\cite{hajic-etal-2009-conll}.
It represents sentence meanings with predicate-argument structures.
Predicates (\eg ``comforted'' in Figure~\ref{fig:teaser}) are usually evoked by verbs and reflect events (\emph{what happened}). Arguments, usually spans of tokens, realize thematic roles of predicates; they include \emph{core} arguments such as \emph{who} did something (\eg ``the doctor'' in Figure~\ref{fig:teaser}) and \emph{to whom} (``the athlete''), as well as \emph{adjunct} arguments like \emph{where} something happened (``In the operation room'') and \emph{how}.

\textbf{Keywords} 
for steering the actual generated content of predicates and arguments.
The keywords can be \revised{\emph{complete} and fully specify the target text of a given span (\eg ``the doctor'' for the \optag{AGENT} in Table~\ref{tab:example-prompts}A), \textit{sparse} and add no constraints beyond the semantic role (\eg * for \optag{LOCATIVE}), or \textit{partial} and specify some of the target text (\eg ``athlete'' for \optag{PATIENT}).
}
As later shown in Table~\ref{table:optag}, these keyword controls are important for supporting a variety of perturbation strategies and applications. 

\textbf{Span ordering} for determining how the thematic roles should be combined.
We use predicate form to control the order of core arguments. 
For example, to distinguish ``the athlete was comforted by the doctor'' from the semantically equivalent ``the doctor comforted the athlete,'' we target the former ordering through a \textit{passive} control, and the latter through an \textit{active} control. Additionally, we use the location of blank tokens (\tblank{*} in Figure~\ref{fig:teaser} and Table~\ref{tab:example-prompts}) to determine the position of generated arguments~\cite{polyjuice:acl21} --- \eg where ``in the operating room'' appears in the generation.

\subsection{Input Format Design}
\label{ssec:prompts}
\begin{table}
\centering
\small
\fontsize{8.5}{9.5}\selectfont
\begin{tabular}{p{0.95\linewidth} @{} }
\toprule
\textbf{Predicate} control: \theaderVerb{active}{past}{comfort}\\
\midrule
\textbf{Primary predicate label} (Always \optag{VERB}) \\
\textbf{Lemma} (Any verb lemma) \\
\textbf{Voice} (\optag{active}, \optag{passive})\footnotemark{} \\
\textbf{Tense} (\optag{past}, \optag{present}, \optag{future}) \\
\midrule\midrule

\textbf{Argument} control: \theader{PATIENT}{partial}{athlete}\\
\midrule
\textbf{Primary argument label} (\optag{AGENT}, \optag{PATIENT}, \optag{TEMPORAL}, \optag{LOCATIVE}, \optag{MANNER}, \optag{CAUSE}, 
    \optag{EXTENT},
    \optag{PURPOSE}, 
    etc.) \\
\textbf{Keyword Content} (\optag{*} symbol or any text) \\
\textbf{Keyword Specificity} (\optag{complete}, \optag{partial}, \optag{sparse}) \\ 
\bottomrule

\end{tabular}
\vspace{-5pt}
\caption{\sysname's \textbf{control codes}.
{Primary} controls build on predicate/argument labels,
and others affect the form and content of generations (More in \sect{ssec:appendix-prompt-format}).
}
\vspace{-10pt}
\label{tab:controls}
\end{table}
\footnotetext{We use \url{http://spacy.io/} for verb or POS detection.}

We integrate the aforementioned controls into the input format detailed in \sect{ssec:appendix-prompt-format} and finetune seq2seq models to output corresponding full sentences.

As shown in Table~\ref{tab:example-prompts}, 
we start our input with a bracketed \textbf{header}, which contains a series of abstract \emph{control codes} (Table~\ref{tab:controls}) that denote the semantic role and keywords (content/specificity) \revised{to realize for each predicate and argument}. \revised{For example, in Table~\ref{tab:example-prompts}A, the control codes for the predicate are ``\textcolor{cop}{\theader{VERB}{active}{past}}'' and the agent argument are ``\textcolor{cop}{\theader{AGENT}{complete}{the doctor}}.''}
We map original semantic roles in PropBank to human-readable labels (\ie \optag{ARG0} $\rightarrow$ \optag{AGENT}) in order to leverage knowledge learned by pretrained models about roles' meanings \citep{paolini2021structured}.

After the header, we append the \textbf{context}, which consists of text to preserved and \emph{blanks} specifying where new text should be generated.
Given such inputs, we train our generator to output text augmented with control codes and brackets, which together specify which generated spans correspond to which controls.
For example, in Table~\ref{tab:example-prompts}B, ``[\tlabel{LOCATIVE}: In the operating room]'' represents the target span of control codes ``\textcolor{cop}{\theader{LOCATIVE}{sparse}{*}}'' and is generated at the location of \emph{blank} \texttt{<id\_0>} right before the preserved \emph{context} ``the doctor.''

\revised{We make three key design choices to allow \sysname to generate roles fluently even when the optimal ordering of roles is unknown (\eg when introducing a new argument).
First, we explicitly separate signal about role placement (\eg blanks in the context) from the role's semantic controls (\eg control codes in the header) such that we can specify the target semantic attributes for a role without tying them to a specific target placement.}
Second, we order the control codes in the header in an input-independent way (see \sect{ssec:appendix-prompt-format}) to discourage the generator from learning to rely on their relative orders.
Third, we insert extra empty blanks into the context (\eg \tblank{3} in Table~\ref{tab:example-prompts}B) such that the \sysname generator can generate spans in the blank locations that result in the most fluent text.

With this flexibility in argument ordering comes the challenge of making strict controls on a single argument: Even if we only want to change verb tense, the generator may reorder other arguments.
\revised{To enable
strict control over generations}, which facilitates minimal perturbations~\cite{ross2020explaining}, we further vary the number of arguments encoded in the header.
As in Table~\ref{tab:example-prompts}C, our generator can take inputs that only mask a subset of arguments, such that, \eg any changes on the \optag{LOCATIVE} argument or \optag{VERB} do not affect the agent and patient. 

\newcommand{\subrule}{\arrayrulecolor{white!30}\specialrule{0pt}{1.5pt}{1.5pt}}
\newcommand{\mainrule}{\arrayrulecolor{black}\midrule}

\renewcommand{\arraystretch}{0.9}
\begin{table*}[h]
\fontsize{8}{9}\selectfont

\begin{subtable}[h]{ 0.44\textwidth}
\centering
\begin{tabular}{@{} p{0.11\textwidth} | p{0.88\textwidth} @{}}
\toprule
\multicolumn{2}{l}{\textbf{(a) Syntactically controlled rewriting}}\\
\midrule\midrule
	Strategy & 
	    \optag{CHANGE\_VTENSE(present)} \newline $\veryshortarrow$
	    [\theaderVerb{active}{\swap{past}{present}}{comfort}] \\
	\subrule
	Perturb. & 
	    In the operation room, the doctor \add{comforts} the athlete.\\

\mainrule
	Strategy &
	\optag{CHANGE\_VVOICE(passive)} \newline $\veryshortarrow$
	[\theaderVerb{\swap{active}{passive}}{past}{comfort}] \\
	\subrule
	Perturb. & 
	    In...room, \add{the athlete was comforted by the doctor}.\\
\mainrule
	Strategy & 
	    \optag{CHANGE\_IDX(4:0)} \newline $\veryshortarrow$
	    \add{\tblank{0}} In the operation room \remove{\tblank{0}}\\
	\subrule
	Perturb. & 
	    \add{{The doctor comforted the athlete}} in the operation room.\\
\mainrule
	Strategy & 
	    \optag{CORE(SWAP\_CORE)} \newline $\veryshortarrow$ [\theader{AGENT}{complete}{the \swap{athlete}{doctor}\newline
        | \theader{PATIENT}{complete}{the \swap{doctor}{athlete}} ]}\\
	\subrule
	Perturb. & 
	In the operation room, \add{the athlete} comforted \add{the doctor}.\\
\arrayrulecolor{black!100}\bottomrule
\end{tabular}
\end{subtable}
\hspace{22pt}
\begin{subtable}[h]{0.47\textwidth}
\centering
\begin{tabular}{@{} p{0.1\textwidth} | p{0.9\textwidth} @{}}
\toprule
\multicolumn{2}{l}{\textbf{(b) Sentence expansion and abstraction}}\\
	\midrule\midrule
	Strategy & 
	    \optag{LOCATIVE:CHANGE\_SPEC(partial)} \newline $\veryshortarrow$
    	[\theader{LOCATIVE}{\swap{complete}{partial}}{in the operation room}]\\
	\subrule
	Perturb. & 
    	\add{Under the dim light} in the operation room, the doctor comforted the athlete.
	\\
	\mainrule
	Strategy & 
	    \optag{LOCATIVE:DELETE} \newline $\veryshortarrow$
	    \removesout{{[\theader{LOCATIVE}{complete}{in the {operation} room}]}}\\
	\subrule
	Perturb. & 
	    \removesout{In the operation room,} the doctor comforted the athlete.
	\\
	
\arrayrulecolor{black!100}\specialrule{.8pt}{2pt}{9pt}
\arrayrulecolor{black!100}\toprule
\multicolumn{2}{l}{\textbf{(c) Data recombination (with external labels and/or contents)}}\\
	\midrule\midrule
	Strategy & 
    	\optag{CAUSE:CHANGE\_CONTENT(because he was in pain)} \newline $\veryshortarrow$[\theader{CAUSE}{complete}{\add{because he was in pain}}]\\
	\subrule
	Perturb. & 
    	In the operation room the doctor comforted the athlete \add{because he was in pain}.\\
\arrayrulecolor{black!100}\bottomrule
\end{tabular}
\end{subtable}

\vspace{-5pt}
\caption{We design a list of primitive operations on input controls to guide perturbations with the \sysname generator.
}
\vspace{-10pt}
\label{table:optag}
\end{table*}

\subsection{Training}
\label{ssec:training}
We finetune \textsc{T5-base}~\citep{2020t5} on input-output pairs derived from gold semantic roles in OntoNotes 5.0 train~(Table \ref{tab:example-prompts}; \citealp{pradhan-etal-2013-towards}).\footnote{On par with \tfive, the blanks are in the form of \texttt{<extra\_id\_*>}; we refer them as \texttt{<id\_*>} for simplicity.}
To train our generator to handle the different input formats described in \sect{ssec:prompts}, 
for each original input, 
we randomly sample the numbers of arguments to mask, number and placement of extra empty blanks,
and keyword content/specificity for each role. See \S\ref{ssec:training-details} for details.

Standard maximum likelihood estimation (MLE) is insufficient for training our generator to follow the controls,
as there may exist signals beyond the given controls for the form of a generation. 
Consider the input: \textcolor{cop}{[\theaderVerb{{active}}{{past}}{comfort} | \theader{{AGENT}}{{partial}}{athlete} | \theader{{PATIENT}}{{complete}}{the doctor}] In the operating room, \tblank{0}, \tblank{1} \tblank{2}.} A generator trained with MLE may ignore controls \optag{AGENT} and \optag{PATIENT} and instead output text ``The doctor comforted the athlete'' rather than ``The athlete comforted the doctor,'' as the \revised{training data distribution may reflect that the} former is more natural given context ``in the operation room.''

To encourage reliance on controls, we incorporate \textbf{unlikelihood training}~\citep{Welleck2020Neural} to penalize generations that conflict with input controls. 
That is, besides Table~\ref{tab:example-prompts}A--C which are used for MLE, we also create ``negative'' samples by randomly perturbing the control codes in our header (as in Table~\ref{tab:example-prompts}N, last row), such that most spans in the target output are not aligned with the control codes.
We create up to three negative samples per input by randomly perturbing 1) verb voice/tense and primary controls for arguments, 2) keyword contents, and 3) keyword specificities (\S\ref{ssec:appendix-prompt-format}).{Our final training data consists of 223K positive and 541K negative examples.}

\section{Creating Perturbations with \sysname}
\label{sec:perturbations}

With \sysname, we can create diverse perturbations by 
modifying input controls.
Given an original sentence, we transform it to an input for \sysname by extracting its semantic parses,\footnote{External semantic role labelers can be used when gold annotations are not available. Our experiments use the opensourced implementation of \citet{shi2019simple}: \url{demo.allennlp.org/semantic-role-labeling}, with a test F1 of 86.5 on the Ontonotes 5.0 dataset~\citep{pradhan-etal-2013-towards}.} 
masking spans we wish to modify,
and providing their control codes.
Then, we modify the {control codes} in the input to generate perturbed sentences with \sysname, filtering out degenerate ones.

\paragraph{Primitive perturbation operations.}
We provide an easily-extendable set of perturbation macros, which capture three common types of perturbations in 
prior work, shown in Table~\ref{table:optag}:
First, \emph{syntactic rewriting} primarily involves shuffling text to create paraphrases~\cite{zhang-etal-2019-paws} or adversarial examples~\cite{iyyer2018adversarial}. We implement such shuffling through operations that perturb predicate forms, move blank tokens, and swap keyword contents of arguments. 
Second, \emph{expansion and abstraction} add or remove text fragments from a sentence 
~\cite{polyjuice:acl21}.
We recreate these through operations on keywords (\eg deletion).
Finally, \emph{data recombination} involves recombining existing textual fragments, within or across inputs~\citep{akyurek2020learning, andreas-2020-good}. 
With \optag{CHANGE\_CONTENT}, we can integrate additional context (\eg from corresponding paragraphs in question answering tasks) into generations.

While our control codes are mostly derived from semantic roles, these primitive operations broadly cover both syntactic and semantic changes.
They can also be used in conjunction with external knowledge bases to achieve targeted edits.\footnote{
For example, if combined with WordNet~\cite{miller1998wordnet}, \sysname perturbations may be able to \revised{incorporate a subset of natural logic~\cite{maccartney2014natural}:} 
In Figure~\ref{fig:teaser}, we can create an entailment relationship by replacing  \remove{doctor} with its hyponym \add{adult}.
}, or be composed to achieve more complex perturbation strategies
as shown in \sect{sec:contrast_set}, \sect{sec:data_augment}, and Appendix \sect{sec:appendix-style-transfer}.

\paragraph{Filtering generations.}
We notice that the \sysname generator produces degenerate outputs for some inputs; we exclude these heuristically based on content and perplexity scores (see \sect{sec:degenerations} for details).

\section{Intrinsic Evaluation}
\label{sec:evaluation}

\renewcommand{\arraystretch}{0.9}

\begin{table*}[tb]
\fontsize{8.5}{9.5}\selectfont
\centering

\begin{tabular}{@{} l  ccc m{0.001em} ccc  m{0.001em} ccc @{}}
\toprule
    & \multicolumn{3}{c}{\bf Closeness}
    && \multicolumn{3}{c}{\bf Pred. Controllability}
    && \multicolumn{3}{c}{\bf Arg. Controllability} \\
\cmidrule(lr){2-4}
\cmidrule(lr){6-8}
\cmidrule(lr){10-12}
{\bf Generator}
    & $F$1        & Precision     & Recall
    && Lemma     & Tense         & Voice
    && Role      & Content       & Spec.\\
\midrule\midrule
\sysname
    & \textbf{64.3}   & \textbf{66.5}    & \textbf{73.4}
    && \textbf{74.3}   & \textbf{80.3}    & \textbf{81.6}
    && \textbf{70.5}    & \textbf{64.5}   & \textbf{64.5} \\
{\sysname}$_{\text{MLE}}$ 
    & 58.5   & 59.5    & 68.6
    && 72.2   & 70.2    & 76.1
    && 60.3    & 45.1   & 45.1 \\
\bottomrule
\end{tabular}
    \vspace{-5pt}
    \caption{
    Intrinsic evaluation performance in percentage.
    \sysname generates perturbations that are close to the original sentence, while reasonably following all the controls specified in Table~\ref{tab:controls}.
    Ablating unlikelihood training ({\sysname}$_{\text{MLE}}$) hurts all metrics across the board. }
    \vspace{-15pt}
    \label{table:intrinsic}
\end{table*}

Following previous work
~\cite{polyjuice:acl21, ross2020explaining}, we evaluate \sysname generations on sentence likelihood, controllability, and closeness.\footnote{We omit the diversity evaluation in \polyjuice, as the keyword content control inherently impacts lexical diversity.}
We additionally evaluate \sysname's unique ability to make fine-grained and compositional perturbations.

\paragraph{Metrics.}
\emph{Likelihood} measures whether the generated text is grammatically correct and semantically meaningful.
Following \citet{ross2020explaining}, we ask whether perturbing a sentence with \sysname drastically changes its likelihood. 
Using a pretrained \gptbaseline, we compute \revised{language modeling} losses for both the original and edited texts and report the ratio of edited / original.
We desire a value of 1.0, which indicates equivalent losses for the two.

\emph{Controllability} measures if the generator responds to the
controls given in inputs.
We rely on cycle consistency to evaluate the controls in Table~\ref{tab:controls}: \revised{For a given generation, we check whether the predicted semantic roles from an SRL system match the control codes in the input} (\eg whether ``in the midst of the earthquake'' in Figure~\ref{fig:teaser} gets detected with a \optag{TEMPORAL} tag).
Since SRL predictions can be noisy, we manually inspect a subset of 98 generated spans and verify that cycle consistency measures positively correlate with ground-truth controllability, with Matthews correlation coefficient $\phi=0.49$ (more details in \S\ref{sec:appendix-intrinsic-eval}).

\emph{Closeness} captures whether the generated sentence involves only necessary changes.
Since our generator takes controls at the argument level, we measure closeness with a weighted F1 score on the expected-to-change and actually-changed spans in the original sentence.
We identify expected-to-change spans from perturbation operations; 
in Figure~\ref{fig:teaser}A, all spans should be changed except for agent ``the doctor.''
Then, we deem a span actually edited if $\geq50\%$ tokens within a span are changed (\eg ``operation room'' in \optag{LOCATIVE}).\footnote{\revised{We empirically tune the threshold to be 50\%, as it tolerates cases where we do not know exactly how the tokens should change (\eg when changing keyword sparsity, we do not know exactly how many new tokens should be generated; when changing semantic role controls, we may want to allow some tokens, like particles, to reoccur, while expecting others in the span to change.)}}
We weigh spans by their lengths to arrive at the final F1 score.

\emph{Compositionality.}
We evaluate \sysname without any finetuning on the \textsc{StylePTB} benchmark~\citep{Lyu2021StylePTBAC}, which builds on the Penn Treebank and assesses both \textit{single}, fine-grained transfers (\eg \textit{To Future Tense}) and \textit{compositional} ones that concurrently edit multiple dimensions (\eg \textit{To Future Tense+ Active To Passive}).
We report mean BLEU scores and compare to the transfer-specific baselines reported in the \textsc{StylePTB} paper (See \sect{sec:appendix-style-transfer}).

\paragraph{Data.} 
We use \textsc{StylePTB} \citep{Lyu2021StylePTBAC} to evaluate compositionality. For other metrics, we perturb 1,000 randomly selected sentences from the OntoNotes 5.0 validation dataset, created the same way as negative samples during training (\S\ref{ssec:appendix-prompt-format}), and evaluate on these perturbations.\footnote{Because these perturbations are generated randomly, some result in sets of controls that are \textit{impossible} to follow. 
Thus, these results represent a lower bound on \sysname's controllability in downstream applications, for which strategies would be designed in a more principled, targeted manner, restricting the perturbations to result in more plausible sets of controls. See \sect{sec:appendix-intrinsic-eval} for more details.}

\subsection{Results}
\sysname generates perturbations with a loss ratio of 0.982, indicating no notable change in language modeling loss after
\revised{perturbation}.
As shown in Table~\ref{table:intrinsic}, \sysname perturbations also tend to be close to the original sentence (F1 $=64.3\%$), with reasonably correct predicates (74.3\%-81.6\% of the time) and arguments (70.5\% controllability on semantic roles and 64.5\% on contents.)
\sysname also demonstrates the ability to make compositional changes; it achieves results comparable to those of fine-tuned baselines on 8/9 tested transfers, and even outperforms the fine-tuned baseline on 5 of them (See \S\ref{sec:appendix-style-transfer} and Table~\ref{tab:styleptb_full} for more details).

\paragraph{Effect of Unlikelihood Training.}
We compare \sysname with a baseline that is finetuned on \tfive \emph{without} unlikelihood training (called {\sysname}$_{\text{MLE}}$ in Table~\ref{table:intrinsic}). Across all metrics, unlikelihood training outperforms {\sysname}$_{\text{MLE}}$, with more controllable and 
\revised{closer}
perturbations (up to a 20\% increase).

\paragraph{Modulating likelihood and closeness.}
As mentioned in \S\ref{ssec:prompts}, our input format supports modulating likelihood and closeness. 
We can increase closeness by only masking the arguments we want to perturb. 
To quantify this effect, we randomly select a single argument to perturb for 1K sentences, but vary the number of masked arguments and number of inserted blanks. 
\revised{As desired,} closeness is maximized
when we mask only the argument we wish to perturb, as in Table~\ref{tab:example-prompts}B (with $F1=67.4\%$), whereas masking two extra arguments and inserting six extra blanks decreases closeness by 3\% and 6\%, respectively.
On the other hand, we can prioritize likelihood (at the cost of closeness) by adding more blanks (\eg insert extra roles whose optimal locations are not known in advance).
On another 1K sentences, we observe that adding six extra blanks increases the likelihood ratio from 0.93 to 0.95.

\newcommand{\tablefrac}{0.85}
\newcommand{\expcell}[1]{\multicolumn{2}{p{\tablefrac\linewidth}}{#1}}
\newcommand{\tasktitle}[3]{
\multicolumn{2}{l|}{\small{\textbf{#1}}} & \small{#2\% (k=#3)} }
\newcommand{\perturbname}[1]{\multicolumn{3}{l}{\small{Perturbation strategy: #1}}}

\renewcommand{\arraystretch}{0.85}

\begin{table*}
\fontsize{8.5}{9.5}\selectfont
\centering
\begin{tabular}{r | l | r }
\toprule
\multicolumn{2}{p{0.8\linewidth}|}{\textbf{Dataset \& Task}} & \textbf{Top-K validity}\\
\mainrule

    
\mainrule
\tasktitle{BoolQ contrast set \normalfont{\cite{gardner2020evaluating}}}{82}{1}\\
\mainrule

Original & \expcell{
    \textbf{Paragraph:}...\uline{his bride} was revealed...Deadpool also discovers that he has a daughter...from a former flame.\newline
    \textbf{Question:} does \tgenerate{AGENT}{Deadpool} \tgenerate{VERB}{have} \tgenerate{PATIENT}{a kid in the comics}? (\textbf{Answer:} True)}\\
\subrule
Strategy & \expcell{Change entity (\optag{AGENT:CHANGE\_CONTENT(his bride)})}\\
\subrule
Perturb. & 
\expcell{\textbf{Question:} does \tgenerate{AGENT}{\add{his bride}} \tgenerate{VERB}{have} \tgenerate{PATIENT}{a kid in the comics}? (\textbf{Answer:} False)}\\

\mainrule
\tasktitle{UD parsing contrast set \normalfont{\cite{gardner2020evaluating}}}{{65}}{10}\\

\mainrule


Original & \expcell{
        
    \textbf{Sentence:} \tgenerate{AGENT}{It} \tgenerate{VERB}{has} \tgenerate{PATIENT}{a diverse range of food at all prices and styles}.\newline
    \textbf{PP attachment}: Noun (``at all prices and styles'' attaches to ``food'')}\\
\subrule
Strategy & \expcell{Swap attachment from noun to verb (\emph{noun$\rightarrow$verb})\newline
    \optag{PATIENT:CHANGE\_CONTENT(a diverse range of food)}
    \newline
    \optag{LOCATIVE:CHANGE\_CONTENT(at),CHANGE\_SPEC(partial)}
    }
    \\
\subrule
Perturb. & 
\expcell{
    \textbf{Sentence:} \tgenerate{AGENT}{It} \tgenerate{VERB}{has} \tgenerate{PATIENT}{a diverse range of food} \tgenerate{LOCATIVE}{\add{at every turn}}.\newline
    \textbf{PP attachment}: Verb (``at every turn'' attaches to ``has'')}\\

\mainrule
\tasktitle{MATRES contrast set \normalfont{\cite{gardner2020evaluating}} }{71}{1}\\
\mainrule
\tasktitle{QA implication \normalfont{\cite{ribeiro2019red}}}{81}{1}\\

\arrayrulecolor{black}\bottomrule
\end{tabular}
\vspace{-5pt}
\caption{A demonstration of how we recreate contrast sets. Using primitive operations in Table~\ref{table:optag}, \sysname supports context-aware and compositional changes. More examples (\eg changing PP attachment \emph{noun$\rightarrow$verb}) are in \S\ref{sec:appendix-contrast-set}.}
\vspace{-10pt}
\label{table:challenge_set_table}
\end{table*}

\section{Contrast Set Creation}
\label{sec:contrast_set}

Manually creating contrast sets is expensive, \eg \citet{gardner2020evaluating} reported spending 10-15 minutes per perturbation for UD Parsing,
whereas labeling existing data is more efficient~\cite{polyjuice:acl21}. 
We show that \sysname can reduce human labor by automatically generating contrast set instances such that annotators only have to label them. \revised{We create \sysname-generated contrast sets for} four tasks: boolean question answering (BoolQ:~\citealp{clark2019boolq}), extractive QA (SQuAD:~\citealp{rajpurkar2016squad}), dependency tree parsing (UD English:~\citealp{nivre2016universal}), and temporal relation extraction (MATRES:~\citealp{ning-etal-2018-multi}).\footnote{\sysname-generated contrast sets are available at \repourl.}

\subsection{Replicating Contrast Sets with \sysname}
\label{subsec:contrast_set_strategy}
We take advantage of two key properties of \sysname:
First, \sysname can make \textbf{context-dependent} changes.
To recreate the \emph{BoolQ contrast set}, we replicate \emph{Entity Change} in \citet{gardner2020evaluating} by replacing content keywords in questions with words in the paragraph that have the same semantic roles.
For example, the paragraph in Table~\ref{table:challenge_set_table} indicates that ``his bride'' can serve as an \texttt{AGENT}.
Second, \sysname allows for \textbf{compositonal} changes.
For example, as in Table~\ref{table:challenge_set_table}, we change prepositional phrase (PP) attachments from \emph{noun$\rightarrow$verb} to recreate the \emph{UD Parsing contrast set} through the following composition of perturbation operations: remove the prepositional phrase from the patient keyword (\eg ``a diverse range of food \removesout{at all prices and styles}''), and introduce an adjunct argument with the preposition as partial keyword (\eg \optag{LOCATIVE} ``at'').
More details are in \S\ref{subsec:appendix-contrast-set-create}.

\textbf{Contrast set validity.}
We consider our perturbation strategies successful if they help reduce human labor, \ie a contrast set author can easily label or take inspiration from \sysname's generations.
Two authors sampled 100 original instances per task, inspected the \emph{top-K} \sysname perturbations, and labeled an instance to be \textbf{valid} if there is at least one perturbation that changes the groundtruth answer while being fluent or requiring only minor fixes.\footnote{Because we exercised controls at different granularity (\ie UD requires sourcing contents from the generator while others mostly require syntactic rewrites with predetermined content), we set $k=10$ for UD---an upper bound for not overloading the human inspector---and $k=1$ for other tasks.}
Table~\ref{table:challenge_set_table} shows that these \sysname perturbation strategies generate contrast sets with high validity.\footnote{\sysname achieves higher validity changing attachment from \emph{noun$\rightarrow$verb} (82\%) than \emph{verb$\rightarrow$noun} (48\%). Discussion in \sect{sec:appendix-contrast-set}.}

\subsection{Measuring Contrast Set Quality} 

We sanity check that \sysname-generated contrast sets can be used to reveal model errors. 
For example, a \textsc{T5-base} model finetuned on BoolQ (with test accuracy 83\%) has a performance of 65\% on both \sysname-generated contrast sets and \citet{gardner2020evaluating}'s (more in \S\ref{subsec:appendix-contrast-set-eval}).
However, this metric is only a proxy for the quality of evaluation data, since it can be made intentionally low if we generate all examples to target a known model error.
Thus, we directly analyze the quality of \sysname contrast sets by measuring their \textbf{lexical diversity} and impact on \revised{token-level} \textbf{dataset artifacts}, both of which play important roles in dataset debiasing. 



We measure {lexical diversity} on UD Parsing contrast sets because it involves sufficient generation of new content. 
We compare \sysname- and human-generated \citep{gardner2020evaluating} contrastive edits for the same 100 UD instances:
we randomly sample one edit for each valid instance, heuristically extract modified PPs, and compute diversity as the ratio of unique to total new tokens in the PPs, filtering stopwords.
For \emph{noun$\rightarrow$verb},
the ratios are respectively 0.78 and 0.99 for \sysname and humans;
for  \emph{verb$\rightarrow$noun}, both are 1.0.
Thus, \sysname can help generate contrast sets without significantly reducing lexical diversity. 
\revised{Furthermore,} \sysname outputs are distinguishable from humans': their unique tokens only overlap for $<15\%$ in \emph{verb$\rightarrow$noun}, and $\sim$6\% for \emph{noun$\rightarrow$verb}, suggesting that \sysname can be used as a collaborative tool to diversify generation.

\begin{figure}[t]
\centering
\includegraphics[width=1\columnwidth]{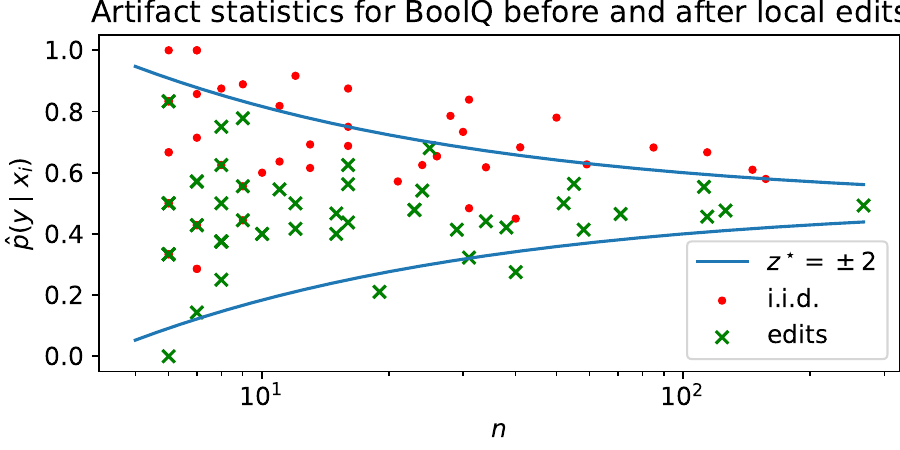}
\vspace{-10pt}
\caption{Dataset artifacts in original BoolQ validation set vs. contrast set created with \sysname \revised{using \cite{gardner2021competency}'s statistical test.
}
}
\vspace{-15pt}
\label{fig:competency}
\end{figure}

We also ask whether \sysname perturbations can reduce {dataset artifacts}.
\revised{
\citet{gardner2021competency} devise a statistical test for dataset artifacts that builds on the argument that no simple feature (\eg single token) should show statistically significant correlation with labels in a language understanding problem. In Figure~\ref{fig:competency}, we display the results:} We plot the numbers of occurrences of each token against the conditional probability of the positive label given that token for both the BoolQ validation data (red dots) and the contrast created by \sysname (green dots). 
All tokens above or below the blue line show statistically significant correlation with positive labels and thus are considered dataset artifacts \revised{in \citet{gardner2021competency}'s framework}. While many tokens in the original BoolQ data 
\revised{exhibit significant correlations},
most in the \sysname contrast set fall within the confidence region. Thus, \sysname can help create less evaluation data \revised{with fewer artifacts}.

\subsection{Discussion}
Across the four tasks, we are able to replicate all perturbation strategies described \revised{by authors of} the original contrast sets.
While \sysname requires manual effort to implement perturbation strategies, we believe the overall saved annotation effort outweighs this initial cost.
First, 
\revised{once implemented, \sysname perturbations can be applied}
to large datasets without requiring additional annotation effort.
\revised{This large-scale applicability} is
especially \revised{useful} for tasks whose single-instance annotation time is significant (\eg UD Parsing).
Second, given that \sysname generations are distinguishable from human ones, they may have the potential to compensate for human omissions and thereby increase test case variety,
which has been shown to be beneficial in prior work~\cite{Ribeiro2020BeyondAB}; \revised{an interesting direction for future work would be to investigate this hypothesis in more detail.} 
Third, the implementation overhead itself diminishes as more strategies are implemented. 
In BoolQ, while \citet{gardner2020evaluating} manually created ``a diverse set of perturbations, including adjective, entity, and event changes'' (see their Appendix B.9), these are all a type of \emph{data recombination} in Table~\ref{table:optag}, and we can unify their implementations with \sysname
into the aforementioned 
keyword replacement in \S\ref{subsec:contrast_set_strategy}.

\section{Data Augmentation}
\label{sec:data_augment}

We explore whether \sysname can be combined with noisy automated labeling for data augmentation.
For the Stanford Natural Language Inference (SNLI) task \citep{bowman2015large}, we show that data augmentation with \sysname perturbations increases model robustness to inference heuristics. 

\citet{min-etal-2020-syntactic} find that augmenting SNLI training data by swapping hypotheses' subject/objects (\eg \textit{This collection contains 16 El Grecos.} $\not\rightarrow$ \textit{16 El Grecos contain this collection}) improves performance on HANS, a challenge set for diagnosing fallible syntactic heuristics in NLI models~\cite{mccoy2019right}. 
Following this, we use \sysname to perturb hypotheses with the \optag{SWAP\_CORE}
operation such that \emph{original hypothesis} $\rightarrow$ \emph{premise} and \emph{perturbed hypothesis} $\rightarrow$ \emph{new hypothesis}.


We finetune \textsc{RoBERTa-base} \citep{liu2019roberta} on different \revised{training datasets}: 
original SNLI train data (unaugmented baseline), 
SNLI train augmented with \citet{min-etal-2020-syntactic} (augmented baseline, referred to as \emph{Syntactic Perturb.} in Table~\ref{tab:hans-results}), 
and SNLI train augmented with \sysname perturbations.
We augment $\sim$2\% of SNLI train.\footnote{We augment the 549,367 SNLI train instances with 10,987 new instances. See \sect{sec:appendix-data-augment} for more details.}
For each subset, we train 20 models with different random seeds.
We evaluate each classifier on the in-domain SNLI test set and the out-of-domain HANS test set.\footnote{For HANS, we follow the standard practice and collapse \emph{neutral} and \emph{contradiction} predictions to \emph{non-entailment}.}


As shown in Table \ref{tab:hans-results}, augmentation with \sysname leads to 5.8-point gain on HANS overall, HANS and a {\textbf{29.2}}-point gain on ``non-entailment,'' compared to the unaugmented baseline.
The improvements are significant, with $t=-6.42$, $p<10^{-3}$ using Student's t-test.
Thus, \sysname perturbations decrease reliance on the lexical-overlap-based inference heuristic for NLI. 

Furthermore, \sysname outperforms \emph{Syntactic Perturb.}, an augmented baseline designed specifically for NLI. We hypothesize that although they create augmentations through similar transformations, 
\citet{min-etal-2020-syntactic}'s approach is limited to inputs with specific syntactic configurations, whereas \sysname's \optag{SWAP\_CORE} argument is applicable to any \optag{AGENT} and \optag{PATIENT} arguments. Thus, \sysname is useful for improving model robustness -- more so than template-based approaches.

\setlength{\tabcolsep}{5pt}

\begin{table}
\centering
\small

\begin{tabular}{@{} l @{\hspace{10pt}} c  @{\hspace{10pt}} c @{\hspace{8pt}} c@{\hspace{8pt}}c @{}}
\toprule

& & \multicolumn{3}{c}{\textbf{HANS Subset}}\\
\cmidrule(lr){3-5}
\textbf{Training Data} & \textbf{SNLI} &  \textbf{All} & \textbf{Entail.} & \textbf{Non-entail.}\\

\midrule\midrule
SNLI Train 
& \textbf{91.1} 
& 64.7 
& \textbf{99.0} 
& 30.5 
\\
+ Syntactic Perturb. 
& {91.0} 
& {67.5} 
& 95.8 
& 39.2 \\ 
+ \textbf{\sysname Perturb.}
& \textbf{91.1} 
& \textbf{70.5} 
& 81.3 
& \textbf{59.7} \\ 

\bottomrule
\end{tabular}
\vspace{-5pt}
\caption{
\sysname augmentations lead to statistically significant gains on the HANS challenge set, without decreasing in-domain accuracy.}
\vspace{-10pt}
\label{tab:hans-results}
\end{table}

\section{Related Work}
\label{sec:related}

Controllable text generation has been widely used to influence various properties of generated text for 
text summarization~\cite{peng2019text}, 
data augmentation~\cite{Lee2021NeuralDA},
style transfer~\cite{reid2021lewis, madaan2020politeness}, 
adversarial example generation~\cite{iyyer2018adversarial}, etc.
Most generators take simple 
\revised{controls}
like 
tense~\cite{hu2017toward},
topic~\cite{Keskar2019CTRLAC}, 
or sentiment polarity~\cite{Dathathri2020PlugAP}, 
which underspecify desired transformations.
\revised{In contrast, \sysname 
concretizes otherwise sparse controls (\eg we can specify making a sentence more negative \emph{through negation}.)} Recent works incorporating syntactic structures for paraphrasing~\cite{iyyer2018adversarial,chen-etal-2019-controllable, bao-etal-2019-generating, kumar2020syntax,sun-etal-2021-aesop,huang2021generating}
or discrete semantic signatures for diverse generation~\cite{weir-etal-2020-cod3s} are similar to \sysname in their high-dimensional specification.




Also closely related are methods that reconstruct sentences from structured semantic representations. \revised{The most similar related work is InFillmore~\cite{ou-etal-2021-infillmore}, which uses semantic representations derived from FrameNet with constrained decoding to guide generation.
While InFillmore tunes the higher-level semantics of a sentence, \sysname's semantic controls incorporate fine-grained information about the location and semantics of textual phrases; in addition, we demonstrate two new applications for semantically-guided generation, contrast set generation and data augmentation.
Abstract Meaning Representation~\cite{banarescu2013abstract, mager2020gpt} is an alternative semantic representation worth exploring \revised{for data perturbation}, as it
may further enable controls on entity recursions~\citep{damonte2019structural}, though expressing such relationships is nontrivial.}

Controlled generators have also been successfully used to perturb text for model training, evaluation, and explanation.
They usually rely on application-specific 
labels~\cite{ross2020explaining, madaan2020generate, sha2021controllingtextedition, akyurek2020learning} or
require pairs of original and perturbed sentences
~\cite{polyjuice:acl21}, which are expensive to generalize. 

Also related are the creation of minimally edited datasets, either through manual rewriting~\cite{gardner2020evaluating, kaushik2019learning}, or creating perturbation templates~\cite{andreas-2020-good, li-etal-2020-linguistically, Ribeiro2020BeyondAB, wu-etal-2019-errudite};
\sysname reduces the human efforts these studies require.
\section{Discussion}
\label{sec:discussion}

We propose \sysname, a 
system that enables task-agnostic, complex and context-aware perturbations.
\revised{\sysname demonstrates that it is possible to drive fine-grained perturbations with semantic features directly derived from an instance.}
\revised{Crucially, it shows that incorporating classical linguistic structures with modern large-scale neural architectures is feasible: With the help of modern pretrained large models, PropBank-style shallow semantic representations can help steer generation towards desired meanings.}


\revised{
\paragraph{Factors that affect \sysname's capability.} 
Though broadly applicable, \sysname's controllability and effectiveness vary for different inputs.
First, creating automatic perturbations with \sysname requires external SRL predictors, which can be noisy on rare semantic roles or low-resource languages.\footnote{\revised{Note that while \sysname is designed to be language agnostic, we only evaluated it on English.}}
Empirically, this did not seem to be a bottleneck, as exposing biases in downstream tasks does not usually require rarity at the semantic role level (\eg testing syntactic heuristics in NLI requires swapping only agents and patients).
However, perturbing more challenging linguistic phenomena may require careful SRL predictor augmentation or even manual semantic role annotation.

We also notice \sysname can sometimes produce degenerate outputs.
We hypothesize that this is a byproduct of unlikelihood training --- \ie the generator learns to reduce the likelihood of negative sequences by generating tokens that are very unlikely to appear in natural text. 
Generation hyperparameters (\eg number of beams) can reduce the number of degenerate outputs.
While we perform unlikelihood training at the sequence level, future work can investigate the effect of penalizing generation at the level of tokens or spans, which may provide finer-grained signals for which spans should be considered unlikely, as well as more strategically balancing positive and negative samples.

\revised{\paragraph{Extending \sysname.} 
We believe the \sysname generator is well-suited for controlled generation tasks beyond the perturbation-based tasks we explore. 
Given key entities or arguments as keywords and fully masked contexts, we envision \sysname can help generate arguments~\cite{schiller2020aspect}, compositionally augment data~\cite{akyurek2020learning}, or generate captions~\cite{chen2020say}.
In particular, as shown in \S\ref{sec:contrast_set}, \sysname's human-readable controls can support humans on data curation, which suggests that designing NLP models for augmenting human capabilities is a promising direction. 

The design of controls is also worthy of in-depth exploration.
As mentioned in \S\ref{sec:related}, AMR might be an alternative for semantic representation, if our primary goal is to 
express non-sequential relations.
On the other hand, dependency parsing labels are useful for syntactic changes; future work may try to balance syntactic and semantic controls.}

Having noted these opportunities, we believe \sysname is already a powerful tool for perturbation, particularly for tasks where compositional changes are required. 
\sysname is opensource, and available at \repourl.}

\section*{Acknowledgements}
We thank 
Ana Marasović, 
William Merrill,
Thomas R. McCoy,
and Daniel S. Weld
for their helpful suggestions, and the anonymous reviewers for their feedback.
Hao Peng is supported by a Google Fellowship.

\bibliography{ref_new}
\bibliographystyle{acl_natbib}

\clearpage
\newpage

\appendix

\begin{appendices}
\section{\sysname Generator Details}
\subsection{Input and Output Formats}
\label{ssec:appendix-prompt-format}

All headers in inputs to the \sysname generator begin with predicate controls, followed by core argument controls (first \optag{AGENT}, then \optag{PATIENT}), and then randomly ordered adjunct argument controls (\optag{LOCATIVE}, \optag{TEMPORAL}, etc.). Secondary controls are always given in the order of \textit{control code+voice+tense:lemma} for verbs and \textit{control code+keyword specificity:keyword content} for arguments. We also blank the auxiliary verbs of the predicate in an input, using \texttt{spacy} to detect them. We exclude discontinuous arguments (\eg those with raw SRL labels \optag{B-C-*}), as well as those with referents (\eg those with raw SRL labels \optag{B-R-*}), from input headers. We map \optag{ARG0} $\rightarrow$ \optag{AGENT} and \optag{ARG1} $\rightarrow$ \optag{PATIENT}. For other numbered arguments, we create human-readable labels by using argument functions included in the PropBank frame for the given predicate \citep{palmer-etal-2005-proposition}.

On the output side, we ask the model to generate the full sentence (Table~\ref{tab:example-prompts}). 
We add the semantic roles for all the generated arguments, to help the generator build explicit mappings between the input control codes and the output spans -- this can be important when the input codes are ambiguous (\eg a \optag{TEMPORAL} argument and a \optag{LOCATIVE} argument that both have keywords ``in''). To use generations in downstream applications, we remove these control codes to obtain cleaned outputs using regular expression matching.

\subsection{Training details}
\label{ssec:training-details}

\paragraph{Training inputs.}
During training, we randomly select, with equal probabilities, whether to mask all arguments or a subset. If a subset, we uniformly select the proportion of arguments to mask. To determine the number of extra blanks, we uniformly select a value less than 10 and set the number of blanks to be the maximum of that selected value and the number of arguments to mask. Any extra blanks (\ie remaining after masking arguments) are inserted between subtrees of the predicate.

We also randomly select keyword contents and keyword specificities. For each argument span, we extract, using \texttt{spacy}, four keyword types from the span: \emph{noun chunks}, \emph{random subtrees}, \emph{exact} keywords, and \emph{prefixes}. For prefixes, we uniformly select a number of tokens to include as the keyword (from 1 to the entire span). Once we extract all keyword candidates, we create corresponding keyword specificities: A keyword is \textit{complete} if it contains all tokens in the original span, \textit{partial} if it contains at least all but 5 tokens, and \textit{sparse} otherwise. Then, we uniformly select a keyword content/specificity pair for each span from the set of keyword candidates (including the \optag{*} symbol).\footnote{Because of how keywords are sampled, we notice that the generator is sensitive to the case of keyword contents. For example, if the keyword for a temporal span is \textit{In 1980} instead of \textit{in 1980}, \sysname is biased towards generating it at the beginning of the sentence. We hypothesize that because some of the keywords we sample during training are cased (\eg \emph{exact} will lead to a cased keyword for a capitalized span beginning a sentence), the generator learns a bias towards generating spans with uppercase keyword at the beginning of the sentence. In applying the generator to perturbations, the case of keyword contents can be used to manipulate the order of generated roles when a certain order of generated contents is desired; otherwise, uncased keywords can be used.}

To generate unlikelihood samples, we use three perturbation strategies on inputs: 1) Change \emph{semantic roles} by swapping thematic role control codes (agent/patient), changing adjunct argument control codes to a uniformly selected other adjunct control code, and changing verb tense/voice. We swap verb tense/voice because the control code \optag{VERB} does not have natural candidate swaps, given that predicates are the building block for semantic parses. We also swap the control codes in the target output. 2) Change keyword \emph{contents} by replacing verb lemmas and keywords for both the predicate and all arguments. To make content swaps, we first gather the most commonly occurring keyword contents for each argument and predicate in Ontonotes 5.0 train, extracted according to the same process as described above for creating training inputs. For each primary control code and keyword specificity (\eg \optag{TEMPORAL+partial}), we store the $15$ most commonly occurring keyword contents. To create the negative inputs, for each span, we uniformly sample from these stored keywords given the span's control code and keyword specificity. This perturbation is designed to discourage the generator from ignoring the keyword content and merely generating commonly occurring text for particular semantic roles. 3) Change keyword \emph{specificities} by uniformly selecting a different specificity. We weight each unlikelihood sample equally, with a reward of -1 (vs +1 for positive samples).

\paragraph{Hyperparameters.} We train the \sysname generator using \texttt{Transformers} \citep{wolf-etal-2020-transformers} for 10 epochs with early stopping. We use batch size 4 and default values for other parameters (learning rate of 5e-5, Adam optimizer).
\section{Intrinsic Evaluation Details}
\label{sec:appendix-intrinsic-eval}


\paragraph{Effectiveness of cycle consistency.} To evaluate to what extent cycle consistency reflects true controllability, we conducted additional manual annotation on role-following. We sampled 25 sentences from the Ontonotes 5.0 development set, transformed them into inputs with varying numbers of masked arguments and blank tokens,
and created up to two perturbed inputs per sentence by randomly replacing their blanked adjunct arguments with other candidate semantic roles (using \optag{CHANGE\_TAG}).
The candidate roles were extracted from the frameset for each predicate verb. We also changed the keyword specificity to \optag{SPARSE}, to make these role swaps more plausible.

We collected \sysname and \sysname$_{MLE}$ generations from both the original and perturbed inputs, and one author manually validated the generated span for each specified argument (98 in total). 
Our annotations were \emph{following} or \emph{not following} the control (\ie the span matches/does not match the designated semantic role), or the set of controls can be \emph{impossible to follow} if the human annotator could not think of any generation that would satisfy the control codes, due to a conflict between the role, keywords, and blank placement.
We then computed the Matthews correlation coefficient (MCC) between the controllability of the role label as measured by the SRL predictor with the gold controllability annotations for the subset of roles without annotation \textit{impossible}. 
The MCCs are 0.49 and 0.51 for \sysname$_{MLE}$ and \sysname, respectively, suggesting that the cycle consistency measures positively correlate with true controllability measures.

Additionally, we measure to what extent the controllability measures from cycle consistency correlate with whether a set of controls is \textit{impossible} to follow. The MCCs are -0.33 for both \sysname and \sysname$_{MLE}$; thus, incorrect role-following as measured by cycle consistency is positively correlated with controls that are impossible to follow. 14/98 instances were manually annotated as having impossible-to-follow controls, suggesting that a nontrivial proportion of the generations for which our intrinsic evaluation measures in \sect{sec:evaluation} found to be unaligned with designated role control codes may be explained by impossible-to-follow controls.

\renewcommand{\arraystretch}{1.05}

\begin{table*}
\footnotesize
\centering
\begin{tabular}{r | l | r }
\toprule
\multicolumn{2}{p{0.8\linewidth}|}{\textbf{Dataset \& Task}} & \textbf{Top-K validity}\\
\mainrule

\mainrule
\tasktitle{MATRES contrast set~\cite{gardner2020evaluating} }{71}{1}\\
\mainrule

Original & \expcell{
    \textbf{Sentence:} Volleyball is a popular sport in the area, and \tgenerate{AGENT}{more than 200 people} would be \tgenerate{VERB}{\uline{watching}} \tgenerate{PATIENT}{the game}, the chief \uline{said}.\newline
    \textbf{Order:} \uline{watching} happens \emph{after} \uline{said}}\\
\subrule
\perturbname{Change tense}\\
Edits & \expcell{
    \optag{VERB:CHANGE\_VFORM(past)}\newline
    $\rightarrow$ [\theaderVerb{active}{\swap{present}{past}}{watch}] 
    Volleyball is...200 people
    \tblank{0} the game, the chief said.}\\
\subrule
Perturbed & 
\expcell{
    \textbf{Sentence:} Volleyball is a popular sport in the area, and \tgenerate{AGENT}{more than 200 people} \tgenerate{VERB}{\add{\uline{watched}}} \tgenerate{PATIENT}{the game}, the chief \uline{said}.\newline
    \textbf{Order:} \uline{watched} happens \emph{before} \uline{said}}\\
\subrule
\perturbname{Change order}\\
Edits & \expcell{
    \optag{PATIENT:MOVE}  \newline
    $\rightarrow$ [\theaderVerb{active}{past}{say} | \theader{AGENT}{complete}{Volleyball...the game}] \remove{\tblank{0}}, the chief said \add{\tblank{0}}.}\\
\subrule
Perturbed & 
\expcell{
    \textbf{Sentence:}\tgenerate{AGENT}{the chief} \tgenerate{VERB}{\uline{said}} \tgenerate{PATIENT}{\add{Volleyball is a popular sport in the area, and more than}\newline
    \add{200 people
    would be \uline{watching} the game}}.\newline
    \textbf{Order:} \uline{said} happens \emph{before} \uline{watch}}\\

\mainrule
\tasktitle{BoolQ contrast set~\cite{gardner2020evaluating}}{82}{1}\\
\mainrule

Original & \expcell{
    \textbf{Paragraph:}...\uline{his bride} was revealed in the webcomic...Deadpool also discovers that he has a daughter by the name of Eleanor, from a former flame of Deadpool named Carmelita.\newline
    \textbf{Q:} does \tgenerate{AGENT}{Deadpool} \tgenerate{VERB}{have} \tgenerate{PATIENT}{a kid in the comics}? (\textbf{A:} True)}\\
%
%
\subrule
\perturbname{Change entity}\\
Edits & \expcell{
    \optag{AGENT:CHANGE\_CONTENT(his bride);}  \newline
    $\rightarrow$ [\theaderVerb{active}{present}{have} | 
    \theader{AGENT}{complete}{\swap{Deadpool}{his bride}}] 
    does \tblank{0} \tblank{1} a kid in the comics?}\\
\subrule
Perturbed & 
\expcell{
    \textbf{Q:} does \tgenerate{AGENT}{\add{his bride}} \tgenerate{VERB}{have} \tgenerate{PATIENT}{a kid in the comics}? (\textbf{A:} False)}\\


\mainrule
\tasktitle{UD parsing contrast set (pp attachment)~\cite{gardner2020evaluating}}{65}{10}\\
\mainrule

Original & \expcell{
    \textbf{Sentence:} 
    Do \tgenerate{AGENT}{you} \tgenerate{VERB}{prefer} \tgenerate{PATIENT}{ham, bacon or sausages} \tgenerate{ADVERBIAL}{with your breakfast}?\newline
    \textbf{PP attachment:} Verb (``with your breakfast'' attaches to ``prefer'')}\\
\subrule

\perturbname{Swap attachment to Noun}\\
Edits & \expcell{
    \optag{PATIENT:CHANGE\_CONTENT(ham, bacon or sausages with),CHANGE\_SPEC(partial)}\newline
    \optag{ADVERBIAL:DELETE} \newline
    $\rightarrow$ [\theaderVerb{active}{present}{prefer} | \theader{PATIENT}{\swap{complete}{partial}}{ham, bacon or sausages \add{with}}\removesout{ | \theader{ADVERBIAL}{complete}{with your breakfast}}] 
    \tblank{0} you \tblank{1} \tblank{2} \tblank{3}?}\\
\subrule
Perturbed & 
\expcell{
    \textbf{Sentence:} Do \tgenerate{AGENT}{you} \tgenerate{VERB}{prefer} \tgenerate{PATIENT}{ham, bacon or sausages \add{with bacon on them}}?\newline
    \textbf{PP attachment}: Noun (``with bacon them'' attaches to ``sausages'')}\\
    
\arrayrulecolor{black!30}\midrule

Original & \expcell{
        
    \textbf{Sentence:} \tgenerate{AGENT}{It} \tgenerate{VERB}{has} \tgenerate{PATIENT}{local boutiques and a diverse range of food at all prices and styles}.\newline
    \textbf{PP attachment}: Noun (``at all prices and styles'' attaches to ``food'')}\\
\subrule

\perturbname{Swap attachment to Verb}\\
Edits & \expcell{
    \optag{PATIENT:CHANGE\_CONTENT(local boutiques and a diverse range of food)}\newline
    \optag{LOCATIVE:CHANGE\_CONTENT(at),CHANGE\_SPEC(partial)}\newline
    $\rightarrow$
    [\theaderVerb{active}{present}{have} | \theader{PATIENT}{complete}{local boutiques and a diverse range of food \removesout{at all prices and styles}} | \add{ \theader{LOCATIVE}{partial}{at}}] 
    \tblank{0} you \tblank{1} \tblank{2} \tblank{3}?}\\
\subrule
Perturbed & 
\expcell{
    \textbf{Sentence:} \tgenerate{AGENT}{It} \tgenerate{VERB}{has} \tgenerate{PATIENT}{local boutiques and a diverse range of food} \tgenerate{LOCATIVE}{\add{at every turn}}.\newline
    \textbf{PP attachment}: Verb (``at every turn'' attaches to ``has'')}\\

\mainrule
\tasktitle{QA implication~\cite{ribeiro2019red}}{81}{1}\\
\mainrule
Original & \expcell{
    \textbf{Q:} 
    \tgenerate{MANNER}{How} did \tgenerate{AGENT}{the Huguenots} \tgenerate{VERB}{defend} \tgenerate{PATIENT}{themselves}?\newline
    \textbf{A:} their own militia}\\
\subrule
\perturbname{Swap answer to be agent}\\
Edits & \expcell{
    \optag{AGENT:CONTENT(who); MANNER:CONTENT(their own militia),SPEC(partial)} \newline
    $\rightarrow$ [\theaderVerb{active}{past}{defend} | 
    \theader{AGENT}{complete}{\swap{the Huguenots}{who}} | \theader{PATIENT}{complete}{themselves} | \theader{MANNER}{\swap{complete}{partial}}{\swap{how}{their own militia}}] 
    \tblank{0} \tblank{1} \tblank{2} \tblank{3}?}\\
\subrule
Perturbed & 
\expcell{
    \textbf{Q:} 
    \tgenerate{AGENT}{\add{Who}} \add{has} \tgenerate{VERB}{defended} \tgenerate{PATIENT}{themselves} \tgenerate{MANNER}{\add{by setting up their own militia}}?\newline
    \textbf{A:} the Huguenots}\\

\arrayrulecolor{black}\bottomrule
\end{tabular}
\caption{A demonstration of how we recreate contrast sets for different tasks (\sect{sec:contrast_set}). Using primitive operations in Table~\ref{table:optag}, \sysname supports context-aware and compositional changes.}
\label{table:challenge_set_table_full}
\end{table*}

\section{Degenerate Outputs}
\label{sec:degenerations}
We observe that \sysname produces degenerate outputs for some inputs, as shown in Table \ref{tab:example-degenerations}.
We hypothesize that this is a byproduct of unlikelihood training: The generator may learn to reduce the likelihood of negative sequences by generating tokens that are very unlikely to appear in natural text. Certain generation hyperparameters, such as the number of beams, can reduce the number of degenerate outputs. While we perform unlikelihood training at the sequence level, future work can investigate the effect of penalizing generation at the level of tokens or spans, which may provide finer-grained signals for which spans should be considered unlikely, as well as more strategically balancing positive and negative samples. 

\begin{table*}
\centering
\fontsize{8.5}{9.5}\selectfont
\setlength{\tabcolsep}{5pt}

\begin{tabular}{@{} p{0.5\linewidth}|p{0.45\linewidth} @{}}
\toprule
\textbf{Input} & \textbf{Degenerate Output}\\
\midrule\midrule

   \textcolor{cop}{[\theaderVerb{passive}{past}{lower} | \theaderrandom{AGENT}{*} | \theader{PATIENT}{partial}{corporate bonds} | \theader{TEMPORAL}{complete}{this year}]} One indication of a growing number of junk defaults , Mr. Asquith says , is that about half of the \$ 3 billion of \tblank{0} that \tblank{1} \tblank{2} \tblank{3} to a default rating \tblank{4} \tblank{5} are junk bonds sold during the market 's big issue years of 1984 through 1986 ..
    &  \multirow{6}{\linewidth}{pastra pastra sanatate sanatate pastraurmatoarele sanatateurmatoarele pastradatorita sanatatedatorita pastracresterea sanatate urmeaza sanatateinformatiile sanatatecresterea pastra urmeaza urmeaza pastrainformatiileinformatiiledatoritadatoritaurmatoarele.}
    \\\midrule
   \multirow{4}{\linewidth}{\textcolor{cop}{[\theaderVerb{active}{present}{visit} | \theaderrandom{AGENT}{*} | \theader{PATIENT}{partial}{Galilee} | \theaderrandom{TEMPORAL}{*}]} \tblank{0} went to \tblank{1} \tblank{2} \tblank{3} .} & AG pastra pastra sanatate sanatate pastraurmatoarele sanatateurmatoareleurmatoarele pastrainformatiile sanatateinformatiileinformatiile pastradatorita sanatatedatoritadatoritaurmatoareledatoritainformatiile dumneavoastra sanatate urmeaza sanatatecresterea\\

\bottomrule

\end{tabular}
\vspace{-5pt}
\caption{Example inputs from the validation set for which the \sysname generator outputs degenerate text.}
\label{tab:example-degenerations}
\end{table*}

\paragraph{Filtering.} To exclude degenerations when using \sysname generations in downstream applications, we employ a combination of heuristics and perplexity-based filtering. As shown by the examples in Table \ref{tab:example-degenerations}, degenerate outputs are easy to detect: We can simply search for whether the output includes ``sanatate.'' We also use cutoffs in perplexity scores computed with \gptbaseline to filter degenerations, as degenerations have significantly lower perplexities than non-degenerate outputs: For generations for 300 randomly sampled validation inputs, the \sysname generator produced generations with a mean perplexity of -346.46 for degenerate outputs (12/300) compared to -86.747 for others. 

\section{Contrast Set Details (\S\ref{sec:contrast_set})}
\label{sec:appendix-contrast-set}

\subsection{Perturbation Strategies}
\label{subsec:appendix-contrast-set-create}

In Table~\ref{table:challenge_set_table_full}, we illustrate our perturbation strategies for creating contrast sets.
Besides BoolQ, already introduced in \S\ref{sec:contrast_set}, the \emph{Matres contrast set}~\citet{gardner2020evaluating} relies on within-sentence context:
As a task that requires detecting and changing the temporal order of two verbs, our perturbations heavily rely on their syntactic relationships.
For example, to change the \emph{appearance order} of verbs in text (as described in ~\cite{gardner2020evaluating}), we would take the parent verb as the base predicate, and \optag{MOVE} the text span containing the child verb.

For \emph{QA implication}~\cite{ribeiro2019red}, we combine \sysname with semantic heuristics: by defining mappings between WH-words and answer types (\eg ``who'' and ``the Huguenots''), we can easily create new questions about different targets.

For \emph{UD English}~\cite{nivre2016universal}, we use constrained decoding~\citep{hokamp-liu-2017-lexically} to prevent generation of the original prepositional phrase. Our strategy for changing prepositional phrase (PP) attachments from \emph{verb$\rightarrow$noun} is similar to that of \emph{noun$\rightarrow$verb}, introduced in \sect{sec:contrast_set}. We use the following composition of perturbation operations: append the preposition to the patient keyword (\eg ``ham or sausages \add{with}''), change patient keyword specificity from \optag{\swap{complete}{partial}} (to generate a new PP attaching to the patient), and delete the argument with original verb attachment (\eg \optag{ADVERBIAL} ``with your breakfast'').

We note that \sysname achieves higher validity changing attachment from \emph{noun$\rightarrow$verb} (82\%) than \emph{verb$\rightarrow$noun} (48\%). This result is expected, as all semantic role labeling arguments attach to verb predicates; thus, introducing controls for an SRL argument (\eg \optag{LOCATIVE} with keyword content ``at'') to generate a preopositional phrase with verb attachment (``at every turn'') reflects the training objective of the generator. On the other hand, our \emph{verb$\rightarrow$noun} strategy involves appending the preposition to the keyword control for an argument, and none of our controls explicitly reflect the target attachment of a prepositional phrase within an argument (\eg keyword controls do not specify whether ``with'' should attach to ``sausages'' vs ``ham''). Furthermore, preposition keywords within an SRL argument do not deterministically lead to noun attachments in our training data--Sometimes a preposition within an argument may reflect verb attachment (\eg in the case of {``Do \tgenerate{AGENT}{you} \tgenerate{VERB}{prefer} \tgenerate{PATIENT}{eating with a fork or eating with a knife}?''}; here, ``eating with a fork or eating with a knife'' is the patient of ``prefer'' but prepositional phrase ``with a fork'' attaches to verb ``eating.'') Because the training objective of our generator does not provide deterministic signal for noun attachment outputs, we do not expect our \emph{verb$\rightarrow$noun} strategy to always result in generations with noun attachment. Our \emph{verb$\rightarrow$noun} strategy is instead intended to \textit{facilitate} the collection of text with noun attachment. Future work can investigate incorporating auxiliary signals about target configurations of keyword contents in outputs (\eg that a preposition should depend on a particular word in the span).



\subsection{Predictor Performance Evaluation}
\label{subsec:appendix-contrast-set-eval}

\begin{table}
\small
\centering
\begin{tabular}{@{} l | r | r r r @{}}
\toprule
\multirow{2}{*}{\textbf{Dataset}} & \multirow{2}{*}{\begin{tabular}{@{}c@{}}\textbf{Task Eval} \\ \textbf{Original}\end{tabular}} & \multicolumn{2}{c}{\textbf{Contrast Set}} \\ & & \textbf{Human} $\downarrow$ & \textbf{Tailor} $\downarrow$ \\ 
\midrule
BoolQ & 82.8 & 64.8 (-17.5) & 64.7 (-17.6) \\
SQuAD & 91.8 & 66.1 (-25.7) & 55.3 (-36.5) \\
MATRES & 70.3 & 49.4 (-20.9) & 42.3 (-28.0) \\
\bottomrule
\end{tabular}
\vspace{-5pt}
\caption{Accuracies of predictors on original task evaluation data and contrasts sets.
The performance drops on contrast sets (vs. original test accuracies), shown in parentheses, are similar for \sysname-generated contrast sets and expert-created sets~\cite{gardner2020evaluating, ribeiro2019red}.
}
\vspace{-5pt}
\label{table:contrast_set_result}
\end{table}

The performances of downstream predictors on original task evaluation data and contrast sets, both \sysname-generated and human-expert-generated, are shown in Table~\ref{table:contrast_set_result}.\footnote{We report accuracy on the test set for MATRES and held-out validation sets for BoolQ and SQuAD, which do not have publicly available test sets.}
For SQuAD, we evaluate a fine-tuned \textsc{RoBERTa}, the most downloaded model hosted on Huggingface,\footnote{\url{https://huggingface.co/deepset/roberta-base-squad2}} and use the QA implication challenge set~\citep{rajpurkar2016squad} as the human contrast set.
Since we could not find readily available predictors for BoolQ and MATRES, we formulate these tasks as a text-to-text task and fine-tune \textsc{T5-base} for 10 epochs; we evaluate the checkpoint with the lowest validation loss.\footnote{For MATRES, we format inputs by surrounding verbs with marker {``<el>''} and {``</el>''} and train the predictor to output the label in natural language, \eg ``Mr. Erdogan has long <el> sought </el> an apology... After that raid An Israeli raid on this ship <el> left </el> nine passengers dead...'' $\rightarrow$ ``before''.}

The drops in predictors' accuracies on the \sysname-generated contrast sets (compared to original test accuracies) show that they can be used to reveal model errors not reflected in original validation data.
However, this result should be interpreted with caution, as it is not directly reflective of dataset quality. For instance, if the contrast data tests one error type or is adversarially constructed to include instances where predictors fail, then lower accuracy does not necessarily mean exposing more model errors. Thus, we treat these performance metrics as secondary to other direct metrics of dataset quality, discussed in \S\ref{sec:contrast_set}, and run this analysis on a small number of contrast set instances as a sanity check. That said, the fact that predictors perform poorly on \sysname-generated contrast sets even without including an adversarial component in our contrast set creation suggests that \sysname can be useful for creating evaluation data to find model errors.
\begin{table*}
\centering
\fontsize{8.5}{9.5}\selectfont
\setlength{\tabcolsep}{5pt}

\begin{tabular}{@{} l | l @{}}
\toprule
\textbf{Premise} & \textbf{\sysname-Generated Hypothesis}\\
\midrule\midrule

A lady in shorts is riding a bike. & A bike is riding a lady in shorts.\\\midrule
A band plays drums in the parade. & Drums are playing a band in the parade.\\\midrule
A young woman eating doritos on mars. & Doritos is eating a young woman on mars\\\midrule
A crowd of people is outside watching a surfer. & A surfer is outside watching a crowd of people.\\\midrule
A lady is holding a viola in the woods. & A viola is holding a lady in the woods.\\\midrule
A girl in striped swimsuit is jumps into the ocean to catch fish & Fish is jumps into the ocean to catch a girl in striped swimsuit\\\midrule
A person is training a choir for the upcoming competition. & For the upcoming competition is training a choir has been person\\\midrule
The photographer gathers the bridal party before the ceremony. & The bridal party is gathering the photographer before the ceremony\\

\bottomrule

\end{tabular}
\vspace{-5pt}
\caption{Examples of augmented data in NLI augmentation experiments (\sect{sec:data_augment}). We use original SNLI hypotheses as premises in the augmented data and use \optag{SWAP\_CORE} with \sysname to generate new hypotheses.}
\vspace{-10pt}
\label{tab:hans-examples}
\end{table*}
\section{Data Augmentation Details (\S\ref{sec:data_augment})}
\label{sec:appendix-data-augment}

\paragraph{Augmented data.} 
To create our augmented data, we filter generations by perplexity scores from \gptbaseline such that we retain $75\%$ of generations. Examples of augmented inputs are shown in Table~\ref{tab:hans-examples}.

\paragraph{Classifiers.}

We train all SNLI classifiers, which build on \textsc{RoBERTa-base} \citep{liu2019roberta}, using \texttt{AllenNLP} \citep{Gardner2017AllenNLP}. We train for 10 epochs using the Adam optimizer with a learning rate of 2e-05 and batch size 32; we use early stopping with a patience of 3.

\section{\sysname's fine-grained and compositional perturbations on \styleptb}
\label{sec:appendix-style-transfer}

\begin{table*}
\footnotesize
\centering

\renewcommand{\arraystretch}{0.85}

\begin{subtable}{1\textwidth}
\centering
\footnotesize
\setlength{\tabcolsep}{7.3pt}

\begin{tabular}{@{}p{0.22\textwidth} cccc c l @{}}
\toprule

\multicolumn{1}{p{0.22\textwidth}}{\multirow{3}{*}{(a) Single transfers}}

& \multicolumn{2}{c}{\textbf{Single Finetune}} & \multicolumn{2}{c}{\textbf{Compos. Finetune}} & \multicolumn{2}{c}{\textbf{No Finetune}}\\

\cmidrule(lr){2-3}
\cmidrule(lr){4-5}
\cmidrule(lr){6-7}

\multicolumn{1}{l}{} 
& \textbf{\gptbaseline} & \textbf{\retrieveedit} & 
\textbf{\csgpttv} & \textbf{\csgpttp} & {\textbf{\sysname}} & \textbf{\sysname, Filtered}\\


\midrule\midrule
To Future Tense & 89.5 & \textbf{89.9} & 72.7 & 81.0 & 87.3 & 88.9, 357/364\\
To Past Tense & 83.6 & \textbf{93.5} & 69.4 & 83.4 & 88.4 & 89.3, 216/218\\
To Present Tense & 75.4 & \textbf{90.9} & 73.3 & 82.6 & 71.0 & 84.7, 175/209\\
ADJ or ADV Removal & 64.7 & \textbf{89.7} & --- & --- & 78.1 & 84.3, 224/243\\
PP Front to Back & 39.8 & 54.1 & --- & --- & \textbf{84.2} & 96.9, 20/23\\
PP Removal & 76.3 & \textbf{79.8} & --- & 76.0 & 71.7 & 85.7, 199/238 \\
Active to Passive & 47.6 & \textbf{68.1} & 47.2 & --- & 55.6 & 77.8, 98/137\\

\bottomrule
\end{tabular}
\label{tab:styleptb-full-single}
\end{subtable}

\vspace{0.2cm}

\setlength{\tabcolsep}{6.5pt}

\begin{subtable}{1\textwidth}
\footnotesize
\centering

\begin{tabular}{@{} llcc c l @{}}
\toprule

\multicolumn{2}{l}{\multirow{3}{*}{(b) Compositional transfers}}
& \multicolumn{1}{c}{\textbf{Compos. Finetune}} & \multicolumn{1}{c}{\textbf{Multi-Single Finetune}} & \multicolumn{2}{c}{\textbf{No Finetune}}\\
\cmidrule(lr){3-3}
\cmidrule(lr){4-4}
\cmidrule(lr){5-6}
\multicolumn{1}{l}{} & \multicolumn{1}{c}{} 
& \multicolumn{1}{c}{\textbf{\csgpt}*}
& \multicolumn{1}{c}{\textbf{\csgptzero}*} & \textbf{\sysname}& \textbf{\sysname, Filtered} \\

\midrule\midrule

\multirow{6}{*}{\begin{tabular}{@{}l@{}}\textbf{Tense +} \\
\textbf{Voice}\end{tabular}} & ToPast+ActiveToPassive & 40.9 & 33.7 & \textbf{66.0} & 66.0, 30/30\\
& ToFuture+ActiveToPassive & \textbf{49.6} & 41.9 & 46.8 & 67.0, 90/131\\
& ToFuture+PassiveToActive & 52.8 & 39.9 & \textbf{68.3} & 68.3, 131/131 \\
& ToPast+PassiveToActive & 47.4 & 36.5 & \textbf{70.2} & 70.2, 65/65 \\
& ToPresent+PassiveToActive & 52.3 & 42.4 & \textbf{69.9} & 69.9, 95/95 \\
& ToPresent+ActiveToPassive & \textbf{50.3} & 44.5   & 31.5 & 61.4, 43/84\\
\midrule

\multirow{3}{*}{\begin{tabular}{@{}l@{}}\textbf{Tense +} \\
\textbf{PPRemoval}\end{tabular}} & ToFuture+PPRemoval & 73.8 & 46.5 & \textbf{74.3} & 79.2, 215/229\\
& ToPast+PPRemoval & \textbf{77.2} & 54.2 & 73.8 & 79.7, 100/108\\
& ToPresent+PPRemoval & \textbf{70.9} & 54.5 & 69.1 & 70.4, 153/156\\

\bottomrule
\end{tabular}
\label{tab:styleptb-full-compositional}
\end{subtable}

\centering
\vspace{-5pt}
\caption{
BLEU scores for single and compositional style transfers in \styleptb.
Baseline results are taken from Tables 14-16 and 19-20 in \citet{Lyu2021StylePTBAC}.
* represents the same type of models finetuned on different subsets of styles, \eg {\csgpt}* in (b) includes \csgpttv, trained on all \textit{Tense+Voice} compositional transfers, and \csgpttp, on \textit{Tenses+PP Removal}.
A single \sysname model helps achieve comparable performance on single transfers compared to finetuned baselines, and is more capable on multiple compositional transfers. }
\label{tab:styleptb_full}
\end{table*}

Here, we show how \sysname can be applied to fine-grained style transfer. 
We 
evaluate \sysname without any finetuning\footnote{This evaluation is zero-shot in spirit, as \sysname is not trained on any paired transfers present in \styleptb. 
However, it is unclear if the test inputs in \styleptb overlap with the Ontonotes 5.0 training data, since the two do share some data points~\citep{van-son-etal-2018-resource}, and \styleptb does not seem to preserve original PTB splits.
This leakage may advantage the external SRL predictor in parsing \styleptb test inputs. 
Still, this advantage should be minor, as the evaluated transfers do not require complex semantic role parsing.} on the \textsc{StylePTB} benchmark \citep{Lyu2021StylePTBAC}, which builds on the Penn Treebank and assesses fine-grained stylistic changes, both on \textit{single} transfers (\eg \textit{To Future Tense}) and compositional ones that concurrently edit multiple stylistic dimensions (\eg \textit{To Future Tense+ Active To Passive}).

\paragraph{Transfers Evaluated.} We evaluate on the transfers in \styleptb for which \citet{Lyu2021StylePTBAC} report results, as their baselines require training separate models for each transfer. Within this subset of transfers, we exclude \textit{PP Back to Front} and \textit{Passive to Active} from evaluation, as they contain $< 5$ test inputs. We also exclude the transfers \textit{Substatement Removal}, \textit{Information Addition}, \textit{Adjective Emphasis}, and \textit{Verb/Action Emphasis}, for which our semantic-role-derived inputs are not well-suited. For example, \textit{Substatement Removal} involves removing substatements that represent ``referring'' and ``situations,'' both of which are technical philosophical concepts that cannot be straightforwardly detected through semantic roles. As another example, \textit{Information Addition} requires adding unordered keyword contents to a sentence (eg \textit{the work force provides the third arm of the alliance}; add keywords: \textit{force black} $\rightarrow$ \textit{the work force provides the third arm of the \textbf{black} alliance \textbf{force}}. While the \sysname generator was only trained with ordered arguments, one could extend the keyword contents to also include unordered target tokens.

\paragraph{Perturbation strategies.} For transfers modifying only verb tense (\eg \textit{To Future Tense}), we mask the verb, modal arguments, and negation arguments, as these are relevant to verb conjugations, and make relevant perturbations on the secondary verb control specifying tense. For transfers modifying verb voice, we mask the verb, agent, and patient. For transfers requiring removal of certain parts of speech (POS)---\ie \textit{ADJ or ADV Removal}, \textit{PP Removal}, and all compositional \textit{Tense + PP Removal} sub-transfers ---we first use \texttt{spacy} to detect such POS, next mask all arguments containing them, and finally perturb the keyword contents to remove the POS for these arguments. For \textit{PP Front to Back}, we mask the argument at the beginning of the original text and implement the change using \optag{CHANGE\_IDX}.

We use cased keywords (\ref{ssec:training-details}) to encourage generations with similarly ordered arguments as the original sentence, except for the \textit{PP Front to Back} transfer, which calls for differently ordered arguments. For transfers modifying verb form only, we set the number of extra blanks to be 2 to allow for generation of helper verbs; for other transfers, we allow for 0 extra blanks to preserve the original order of generated spans.
We decode perturbed sentences greedly using beam search (with beam width 10) and preventing repeated bigrams.

For each transfer, we create perturbations for each predicate in the original input, and report mean BLEU scores.\footnote{We report \texttt{Bleu\_1} from \texttt{nlg-eval} \citep{sharma2017nlgeval}.}
Because this process results in multiple perturbations (one per verb), we choose the one with the lowest perplexity from \gptbaseline to represent the transfer. Unsuccessful transfers, either due to a failure of perturbation strategy (\eg no verbs are found by our SRL predictor) or due to a degenerate output (see \sect{sec:degenerations}), are given a BLEU score of 0.0. 


\paragraph{Baselines.}
We work with baselines reported by \citet{Lyu2021StylePTBAC}: \textbf{{\gptbaseline}} and \textbf{{\retrieveedit}} are the best-performing single-transfer models evaluated but require separate models to be trained for each transfer.
\textbf{{\csgpt}*} are models trained on compositional subsets of data (\eg \emph{Tense+Voice}, detailed in Table~\ref{tab:styleptb_full} caption). \textbf{\csgptzero} are ablations of {\csgpt}* trained only on corresponding individual changes but evaluated on compositional transfers.\footnote{\csgptzero refers to \textsc{CS-GPT-Zero} in \citet{Lyu2021StylePTBAC}.}

\paragraph{Result.}
On compositional transfers, we find that \sysname outperforms the baseline system trained without compositional fine-tuning, \csgptzero, on 8/9 compositions, and even outperforms {\csgpt}* (models with compositional finetuning) on 5 cases.
It also achieves compatible or better results than \gptbaseline and \retrieveedit on single transfers.
Low \sysname performance on some transfers
(\eg \textit{ToFuture+ActiveToPassive}) appears to be driven by unsuccessful transfers, rather than generations that do not follow controls, as indicated by the higher performances on the subset where unsuccessful transfers are removed (\emph{Filtered Test}). 
Importantly, \sysname achieves these gains with a \emph{single model} and \emph{without any transfer-specific finetuning}.

\end{appendices}

\end{document}